\documentclass[10pt,twocolumn,letterpaper]{article}

\usepackage[pagenumbers]{cvpr} 

\usepackage{graphicx}
\usepackage{amsmath}
\usepackage{amssymb}
\usepackage{booktabs}
\usepackage{url}

\usepackage[pagebackref,breaklinks,colorlinks]{hyperref}

\makeatletter
\@namedef{ver@everyshi.sty}{}
\makeatother
\usepackage{graphicx}
\usepackage{pgfplots}
\usepackage{filecontents}
\usepackage{pgfplotstable}
\pgfplotsset{compat=newest}
\usepackage{siunitx}

\usepackage[capitalize]{cleveref}
\crefname{section}{Sec.}{Secs.}
\Crefname{section}{Section}{Sections}
\Crefname{table}{Table}{Tables}
\crefname{table}{Tab.}{Tabs.}

\usepackage{multirow}

\usepackage{enumitem,bigdelim}
\usepackage{lipsum}

\newcommand{\boldparagraph}[1]{\vspace{0em}\noindent{\bf #1} }    

\newcommand{\Ro}{\mathbf{r}}
\newcommand{\Rd}{\Delta\Ro}
\newcommand{\To}{\mathbf{t}}
\newcommand{\Td}{\Delta\To}
\newcommand{\mesh}{\Theta}
\newcommand{\Ren}{\mathcal{R}}
\newcommand{\Mot}{\mathcal{M}}

\newcommand{\Ti}{\tau}
\newcommand{\Fen}{F}
\newcommand{\Men}{M}
\newcommand{\Fi}{\Fen_\Ti}
\newcommand{\Mi}{\Men_\Ti}
\newcommand{\Hi}{H_{\Ti}}

\newcommand{\Ir}{\hat{I}}
\newcommand{\Loss}{\mathcal{L}}

\newcommand{\error}{e}
\newcommand{\expos}{\epsilon}
\newcommand{\TrajR}{\mathcal{Q}}
\newcommand{\TrajT}{\mathcal{T}}
\newcommand{\motion}{\Omega}

\newcommand{\MakeArrowTwo}{\multicolumn{2}{c}{\upbracefill \hspace*{0.2em}}}
\newcommand{\MakeArrowVar}[1]{\multicolumn{#1}{c}{\upbracefill \hspace*{0.0em}}}

\newcommand\blfootnote[1]{%
  \begingroup
  \renewcommand\thefootnote{}\footnote{#1}%
  \addtocounter{footnote}{-1}%
  \endgroup
}

\def\cvprPaperID{875} 
\def\confName{CVPR}
\def\confYear{2022}

\begin{document}

\title{Motion-from-Blur: \\ 3D Shape and Motion Estimation of Motion-blurred Objects in Videos}

\author{Denys Rozumnyi$^{1,4}$
\hspace{0.5cm}
Martin R. Oswald$^{1,2}$
\hspace{0.5cm}
Vittorio Ferrari$^{3}$
\hspace{0.5cm}
Marc Pollefeys$^{1}$\\[0.0em]
$^{1}$Department of Computer Science, ETH Zurich
\hspace{0.5cm} 
$^{2}$University of Amsterdam\\
$^{3}$Google Research
\hspace{0.5cm}
$^{4}$Czech Technical University in Prague
\\[0.0em]
{\tt\small \{denys.rozumnyi,martin.oswald,marc.pollefeys\}@inf.ethz.ch}
\hspace{0.5cm} 
{\tt\small vittoferrari@google.com}\\
}
\maketitle


\begin{abstract}
We propose a method for jointly estimating the 3D motion, 3D shape, and appearance of highly motion-blurred objects from a video.
To this end, we model the blurred appearance of a fast moving object in a generative fashion by parametrizing its 3D position, rotation, velocity, acceleration, bounces, shape, and texture over the duration of a predefined time window spanning multiple frames.
Using differentiable rendering, we are able to estimate all parameters by minimizing the pixel-wise reprojection error to the input video via backpropagating through a rendering pipeline that accounts for motion blur by averaging the graphics output over short time intervals.
For that purpose, we also estimate the camera exposure gap time within the same optimization.
To account for abrupt motion changes like bounces, we model the motion trajectory as a piece-wise polynomial, and we are able to estimate the specific time of the bounce at sub-frame accuracy.
Experiments on established benchmark datasets demonstrate that our method outperforms previous methods for fast moving object deblurring and 3D reconstruction. 
\end{abstract}

\section{Introduction} \label{sec:intro}
%
3D object reconstruction from 2D images is one of the key tasks in computer vision~\cite{sfm,nerf,schoenberger2016sfm,schoenberger2016mvs}. 
It allows better modeling of the underlying 3D world.
Applications of 3D object reconstruction are broad, ranging from robotic mapping~\cite{han2021reconstructing} to augmented reality~\cite{Xu_2018_CVPR_Workshops}.
%
Even though some recent methods deal with the extreme and under-constrained case of reconstructing 3D objects from a single 2D image~\cite{pixelnerf,what3d_cvpr19}, most methods take advantage of a multi-view setting~\cite{nerf,sfm,schoenberger2016sfm,schoenberger2016mvs}.
However, all generic 3D object reconstruction methods assume that the object moves slowly compared to the camera frame rate, resulting in sharp 2D images.
%
The task of 3D object reconstruction becomes much more challenging when the object moves fast during the camera exposure time, resulting in a motion-blurred 2D image. 
The Shape-from-Blur (SfB) method~\cite{sfb} tackled this challenging scenario to extract 3D shape and motion from a single motion-blurred image of the object.
This scenario is difficult because motion blur makes the input image noisier, and many high-frequency details are lost.
On the other hand, even a single image gives potentially several views of the object, which are averaged by motion blur into one frame. 
SfB~\cite{sfb} explicitly modeled this phenomenon and successfully exploited it.

\newcommand{\addimgT}[1]{\includegraphics[width=0.12\linewidth]{#1}}
\newcommand{\addimgNov}[1]{\includegraphics[width=0.22\linewidth]{#1}}

\newcommand{\makeOneRow}[1]{\addimgT{#1_mfb0} & \addimgT{#1_mfb7}}
\newcommand{\makeBracket}[1]{\raisebox{3.2em}{\rdelim\}{#1}{*}[]}}

\newcommand{\addNovel}[1]{\raisebox{4.15em}{\multirow{5}{*}{\includegraphics[height=0.88\linewidth]{imgs/key_novel/snapshot0#1}}}}
\newcommand{\addNovelt}[1]{\raisebox{0.01em}{\multirow{5}{*}{\includegraphics[height=0.5\linewidth]{imgs/key_novel/snapshot0#1}}}}

\newcommand{\makeRowT}[1]{%
\addimgT{#1_01_im} & \makeBracket{50} & \raisebox{4.15em}{\multirow{5}{*}{\includegraphics[width=0.223\linewidth]{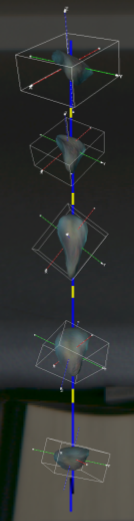}}} & \addNovel{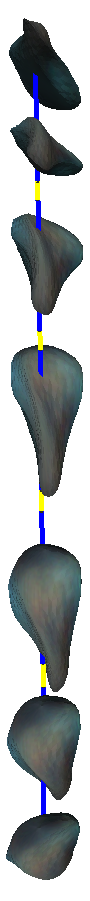} &  \addNovel{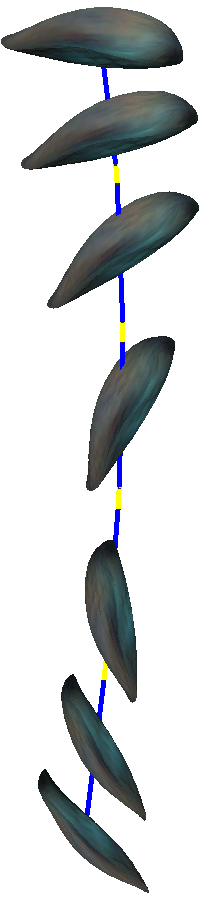}  & \addNovelt{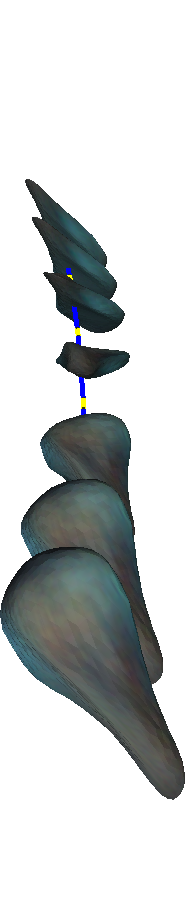}  
& \makeOneRow{#1_01} \\
\addimgT{#1_02_im} & & &  & & & \makeOneRow{#1_02} \\
\addimgT{#1_03_im} & & &  & & & \makeOneRow{#1_03} \\
\addimgT{#1_04_im} & & &  & & & \makeOneRow{#1_04} \\
\addimgT{#1_05_im} & & &  & & & \makeOneRow{#1_05} \\
}

\begin{figure}
\centering
\footnotesize
\setlength{\tabcolsep}{0.0em} 
\renewcommand{\arraystretch}{0.4} 
\begin{tabular}{@{}c@{}c@{}c@{}c@{}c@{}c@{}c@{}ccc@{}}

\makeRowT{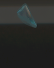}
Input & & Shape \& Motion & \multicolumn{3}{c}{Novel views} & \multicolumn{2}{c}{TSR ($2\times$)} \\ 
\end{tabular}
\vspace{-6pt}
\caption{\textbf{Reconstructing 3D shape and motion of a motion-blurred falling key.} We jointly optimize over multiple input frames to estimate a single 3D textured mesh and corresponding motion model (\textit{blue}: observed trajectory, \textit{yellow}: the exposure gap). Temporal super-resolution (TSR) is one of the applications of the proposed Motion-from-Blur method.}
\label{fig:teaser}
\end{figure}

In this paper, we go beyond previous methods by estimating the 3D object's shape and its motion from a series of motion-blurred video frames.
To achieve this, we optimize all parameters jointly over multiple input frames (\ie the object's 3D shape and texture, as well as its 3D motion).
We tie up the object's 3D shape and texture to be constant over all frames.
Due to the longer time intervals involved, we must model more complex object motions (3D translation and 3D rotation) than necessary for a single motion-blurred frame~\cite{sfb},~\eg the acceleration of a falling object (Fig.~\ref{fig:teaser}), or a ball bouncing against a wall (Fig.~\ref{fig:bounce}).
Using multiple frames also comes with an additional challenge: the camera shutter opens and closes in set time intervals, leading to a gap in the object's visible trajectory and appearance. 
To properly succeed in our task, we must also recover this exposure gap.
For a single frame only (as in \cite{sfb}), the motion direction (forward vs. backward motion along the estimated axis) is ambiguous.
For instance, in Fig.~\ref{fig:teaser}, the key could be translating from top to bottom or vice-versa, both resulting in the same input image.
Since we consider multiple frames jointly, the motion direction is no longer ambiguous and can always be recovered.
Moreover, for rotating objects, we can reconstruct a more complete 3D model as we can integrate more observations covering its total surface. 
In contrast, previous single-frame work~\cite{sfb} produces strong artifacts on unseen parts.
An example of our method's output and an application to temporal super-resolution is shown in Fig.~\ref{fig:teaser}.\\
%
To summarize, we make the following contributions:
\begin{enumerate}[itemsep=0.1pt,topsep=3pt,leftmargin=*,label=\textbf{(\arabic*)}]
    \item We propose a method called Motion-from-Blur (MfB) that jointly estimates the 3D motion, 3D shape, and texture of motion-blurred objects in videos by optimizing over multiple blurred frames. 
    Motion-from-Blur is the first method to optimize over a video sequence instead of a single frame.
    The source code is available at \href{https://github.com/rozumden/MotionFromBlur}{github.com/rozumden/MotionFromBlur}.
    \item Our multi-frame optimization enables the estimation of the motion direction as well as more complex object motions such as acceleration and abrupt direction changes,~\eg bounces, for both 3D translation and 3D rotation.
    Moreover, compared to single-frame approaches, our estimates are also more consistent over time, with always correct motion direction, and more complete 3D shape reconstruction. 
    \item As a requirement to model multiple frames, we estimate the exposure gap as part of the proposed optimization.
\end{enumerate}
%

\section{Related work}  \label{sec:related}
%
Many methods have been proposed for generic deblurring,~\eg~\cite{Chi_2021_CVPR, Li_2021_CVPR, Pan_2020_CVPR, Zhang_2020_CVPR, Suin_2020_CVPR, Kaufman_2020_CVPR, Kupyn_2018_CVPR,Kupyn_2019_ICCV}.
A related task of frame interpolation or temporal super-resolution is studied in~\cite{Gui_2020_CVPR, Niklaus_CVPR_2020, Shen_2020_CVPR, Jin_2019_CVPR, Pan_2020_CVPR, Ding_2021_CVPR, Siyao_2021_CVPR,Jin_2018_CVPR}.
However, none of the generic deblurring methods work on extremely motion-blurred objects as shown in~\cite{fmo}, and specific methods are required.

We focus on deblurring and 3D reconstruction of highly motion-blurred objects.
These are called fast moving objects as defined in~\cite{fmo}~--~objects that move over distances larger than their size within the exposure time of one image.
Detection and tracking of such objects are usually done by classical image processing methods~\cite{fmo,tbd,tbd_ijcv} or more recently by deep learning~\cite{fmodetect,fmo_segmentation}.
A model-based approach to high-speed tracking with a specialized time-of-flight camera is studied in~\cite{Stuhmer_2015_ICCV}.

\boldparagraph{Single-frame deblurring of fast moving objects.}
The first methods for fast moving object deblurring~\cite{fmo, kotera2018} assumed an object with a constant 2D appearance~$F$ and 2D shape mask~$M$. Hence, the object was represented by a single 2D image patch that could only be rigidly translated and rotated in 2D.
They defined the image formation model for such objects as the blending of the blurred object appearance $F$ and the background $B$:
\begin{equation}
	\label{eq:fmo}
	I = H*F + (1-H*M)\, \cdot B \enspace,
\end{equation}
where the motion blur is modeled by the convolution of the sharp object appearance $F$ and its trajectory, defined by the blur kernel~$H$.
Several follow-up methods~\cite{tbd,tbd_ijcv,tbdnc,sroubek2020,tbd3d,kotera2020,fmodetect} were proposed to solve for $(F,M,H)$ given the input image $I$ and background $B$. 
They approximate the solution in a least-squares sense by energy minimization with suitable regularizers summarized by function $\mathop{\mathrm{reg}}(\cdot)$:
\begin{equation}
	\label{eq:fmomin}
	\min_{F,M,H} \frac{1}{2} \| H*F + (1-H*M)\, \cdot B - I \|_2^2 + \mathop{\mathrm{reg}}(F,M,H) \enspace.
\end{equation}	
As common in blind deblurring problems~\cite{kotera2018}, they deploy alternating minimization w.r.t.~object~$(F,M)$ and trajectory~$H$ separately in a loop.
Optimization is made possible thanks to many regularizers such as appearance total variation, blur kernel sparsity~\cite{kotera2018,tbd,tbd_ijcv}, motion blur prior for curves~\cite{sroubek2020}, appearance and mask rotational symmetry~\cite{tbd3d}, among others.
All of these methods share the same drawback that stems from the underlying image formation model~\eqref{eq:fmo}, which assumes a constant 2D object appearance.

TbD-3D~\cite{tbd3d} extended the image formation model to support fast moving objects with a piece-wise constant 2D appearance as
\begin{equation}
	\label{eq:fmo3d}
	I = \sum_{\Ti} \Hi*\Fi + \big(1-\sum_{\Ti} \Hi*\Mi\big)\, \cdot B \enspace,
\end{equation}
where the trajectory is split into several pieces $\Hi$, assuming that along each piece the object appearance $\Fi$ and mask $\Mi$ are constant.
All unknowns are again estimated by energy minimization with additional problem-specific priors,~\eg object appearance in neighboring pieces is similar.

Later, DeFMO~\cite{defmo} was the first learning-based method for fast moving object deblurring, and it generalized the image formation model further to objects with a 2D appearance that can change arbitrarily along the trajectory: 
\begin{equation}
	\label{eq:defmo}
	I = \int_{0}^{1} \Fi \cdot \Mi \:\text{d}\Ti + \Big(1-\int_{0}^{1} \Mi \:\text{d}\Ti \Big)\, \cdot B \enspace,
\end{equation}
where object appearance $\Fi$ and mask $\Mi$ are modeled by an encoder-decoder network.
The network places $(\Fi, \Mi)$ at the right image location, directly encoding the object trajectory.
Although trained on synthetic ShapeNet data~\cite{shapenet2015}, DeFMO was shown to generalize to real-world images.

\begin{figure*}
	\includegraphics[width=\textwidth]{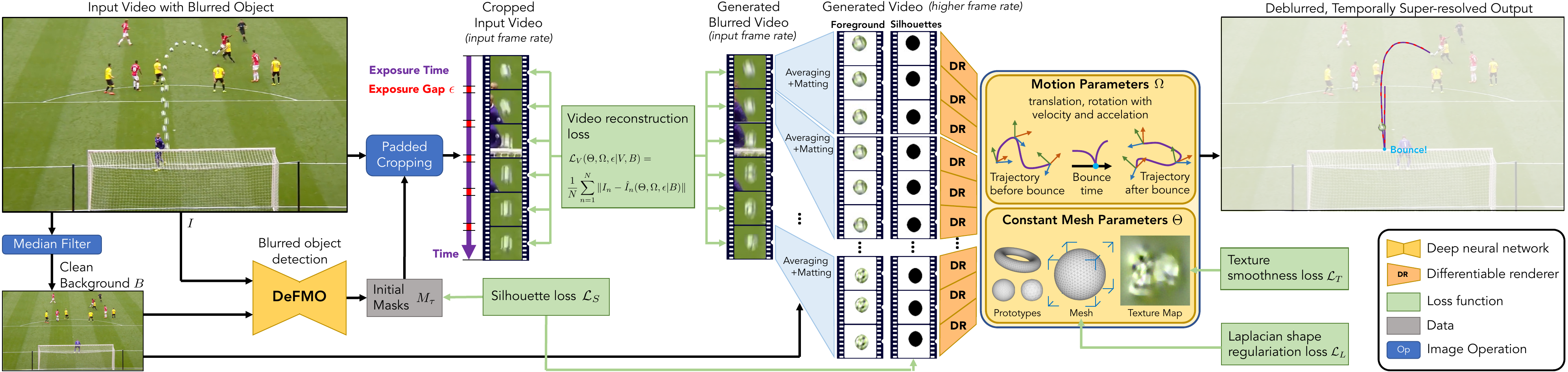}\\[-15pt]
	\caption{\textbf{Overview of Motion-from-Blur (MfB).} For a video of a motion-blurred object, we estimate its 3D motion, 3D shape, and texture. From \textit{right to left}, the pipeline can be interpreted as a generative model: Starting from all parameters for an object and its motion, we render high-frame-rate videos with the object appearance (foreground) and its silhouette. 
	Together with the known background, we generate a motion-blurred video of the object that should match the input video as good as possible.
	At test time, we optimize all object parameters (and the exposure gap) of this inverse problem by backpropagating the image differences through the differentiable renderer~(\textit{left to right}). We initialize the optimization using the DeFMO method~\cite{defmo}, which provides rough silhouettes of the blurred object. 
	MfB models a piece-wise smooth motion path to allow for a motion discontinuity like a bounce.
	Video source: \href{https://youtu.be/wa1ORli2Nyo?t=78}{YouTube}.
	}
	\label{fig:pipeline}
\end{figure*}
%

\boldparagraph{Single-frame 3D reconstruction of fast moving objects.}
The only prior work capable of 3D reconstruction of fast moving objects is Shape-from-Blur~\cite{sfb}.
Instead of merely recovering the 2D object projections~$(\Fi,\Mi)$, they reconstruct the object's 3D shape mesh~$\mesh$ as well as 3D motion.
The latter is represented as the 3D translation $\To$ and 3D rotation $\Ro$, defining the object's pose at the beginning of the exposure time ($\Ti = 0$), and the offsets $\Td$ and $\Rd$, moving the object to its pose at the end of the exposure time ($\Ti = 1$).
With these definitions, the image formation model becomes
\begin{equation}
	\label{eq:sfb}
	\begin{split}
	    I & =  \!\int_{0}^{1} \!\!\Ren_F\big(\Mot(\mesh, \Ro + \Ti \cdot \Rd, \To + \Ti \cdot \Td)\big) \:\text{d}\Ti +  \\
	     & + \Big(1-\!\int_{0}^{1} \!\!\Ren_S\big(\Mot(\mesh, \Ro + \Ti \cdot \Rd, \To + \Ti \cdot \Td)\big) \:\text{d}\Ti \Big)\,\cdot B\enspace,  \\
	\end{split}
\end{equation}
where the function $\Mot$ transforms the mesh $\mesh$ by the given 3D translation and 3D rotation.
Energy minimization is constructed from~\eqref{eq:sfb} to find the mesh and motion parameters that would re-render the input image $I$ as closely as possible.
To make minimization feasible, mesh rendering is made differentiable using Differentialbe Interpolation-Based Rendering~\cite{dibr}, denoted by $\Ren_F$ and $\Ren_S$ for the appearance and 2D object silhouette, respectively.
To differentiate from 2D masks $\Mi$, silhouettes denote real renderings of a 3D object mesh.
In contrast to Shape-from-Blur, our method models more complex trajectories, estimates the exposure gap, and takes into account several frames jointly, thereby allowing temporally consistent predictions and more completely reconstructed 3D shape models.

\boldparagraph{3D shape from sharp images.}
Many methods for 3D reconstruction have been proposed, both for single-frame~\cite{pixelnerf,pixel2mesh,what3d_cvpr19,Richter_2018_CVPR,Fan_2017_CVPR} and multi-frame setting~\cite{nerf,sfm,schoenberger2016sfm,schoenberger2016mvs}. 
But these methods assume sharp objects in the scene (the methods listed in previous paragraphs are the only ones dedicated to fast moving objects).
In other words, they either assume that an object moves slowly compared to the camera frame rate (or, equivalently, that the camera moves slowly).


\section{Method}  \label{sec:method}
%
When images are captured by a conventional camera, the camera opens its shutter to allow the right amount of light to reach the camera sensor.
Then, the shutter closes, and the whole process is repeated until the required number of frames is captured.
This physical reality of the camera capturing process leads to two phenomena, which we model and exploit in our optimization.
The first one is the motion blur that appears when the object moves while the shutter is open. 
The second one is the exposure gap that makes the camera `blind` when the shutter is closed, thus not observing the moving object for some parts of its motion.

We assume the input is a video stream $V = \{I_1, \dots, I_{N}\}$ of $N$ RGB images depicting a fast moving object. 
The desired output of our method is a single textured 3D object mesh $\mesh$, its motion parameters $\motion$ consisting of a continuous 3D translation and 3D rotation at every point in time $\Ti$ during the video duration, and the exposure gap $\expos$ (a real-valued parameter).
Sec.~\ref{sec:model} introduces these parameters and a video formation model to generate video frames for given parameters.
In case we know the real values of all parameters, we could render the input video $V$.
Then, in Sec.~\ref{sec:fitting}, we show how to optimize these parameters to re-render the input video frames as closely as possible.

\subsection{Modeling} \label{sec:model}
\boldparagraph{Mesh modeling.}
The mesh parameters $\mesh$ consist of an index to a prototype mesh, vertex offsets from its initial vertex positions to deform the mesh, and the texture map.
We use a set of prototype meshes to account for varying mesh complexity and different genus numbers.
Our set of prototype meshes contains a torus and two spheres with a different number of vertices. 
The texture mapping from vertices to the 2D location on the texture map is assumed to be fixed.
Similarly, the mesh triangular faces consist of fixed sets of edges that connect vertices.

\boldparagraph{Motion modeling.}
The object motion $\motion$ is composed of continuous 3D translations $\TrajT(\Ti) \in \mathbb{R}^3$ and 3D rotations represented by quaternions $\TrajR(\Ti) \in \mathbb{R}^4$.
Both translations and rotations are viewed from the camera perspective, which is assumed to be static.
We assume that they are defined at all points in time $\Ti \in [0,1]$, spanning the duration of the entire input video.
We implement the functions $\TrajT(\Ti)$ and $\TrajR(\Ti)$ as piece-wise polynomials, and their parameters are the polynomial coefficients.
More precisely, we use piece-wise quadratic functions with two connected pieces, which are able to model one bounce, as well as accelerating motions (\eg a falling object).

\boldparagraph{Exposure modeling.}
We denote the {\em exposure gap} as a real-valued parameter $\expos \in [0,1]$ that represents the fraction of the duration of a frame during which the camera shutter is closed.
In other words, it is the duration of the closed shutter divided by the duration of one shutter cycle.
A hypothetical full exposure camera that never closes its shutter would result in $\expos = 0$.
In most cases, conventional cameras would set their exposure gap $\expos$ close to 0 for dark environments to get as much light as possible and close to 1 for very bright environments to avoid overexposure.
Typically, smaller exposure gaps $\expos$ lead to more motion blur.

\boldparagraph{Video formation model.} 
The video formation model is the core of our method. 
It renders a video frame $\Ir_n$ for a given set of all above-mentioned parameters:
\begin{equation}
	\label{eq:msfb}
	\begin{split}
	    \Ir_n & (\mesh, \motion,\expos | B)  =  \!\int_{\frac{n-1}{N}}^{\frac{n-\expos}{N}} \!\!\Ren_F\bigg(\Mot\Big(\mesh, \TrajR(\Ti), \TrajT(\Ti)\Big)\bigg) \:\text{d}\Ti +  \\
	      & + \Bigg(1-\!\int_{\frac{n-1}{N}}^{\frac{n-\expos}{N}} \!\!\Ren_S\bigg(\Mot\Big(\mesh, \TrajR(\Ti), \TrajT(\Ti)\Big)\bigg) \:\text{d}\Ti \Bigg)\,\cdot B \enspace,  \\
	\end{split}
\end{equation}
where the interval bounds for frame $\Ir_n$ go from the beginning of its exposure time when the shutter opens at time $\Ti = \frac{n-1}{N}$ to the end of its exposure time when the shutter closes at time $\Ti = \frac{n-\expos}{N}$.
Consequently, the object is not observed between $\Ti = \frac{n-\expos}{N}$ and $\Ti = \frac{n}{N}$.
As defined previously, the function $\Mot$ first rotates the mesh $\mesh$ by the 3D rotation $\TrajR(\Ti)$ and then moves it by the 3D translation $\TrajT(\Ti)$.
Mesh rendering is implemented by Differentiable Interpolation-Based Rendering~\cite{dibr}, denoted by $\Ren_F$ for the appearance and by $\Ren_S$ for the silhouette.
Like all previous methods for fast moving object deblurring, we compute the background $B$ as the median of all frames $I_n$ in the input video $V$.
Note that our modeling is a strict generalization of SfB~\cite{sfb} for the case of $N=1$ and linear motion.

\subsection{Model fitting} \label{sec:fitting}
%
This section presents an optimization method to fit the introduced model to the given input video.

\boldparagraph{Loss function.}
The main driving force of the proposed approach is the video reconstruction loss
\begin{equation}
	\label{eq:loss}
    \Loss_V(\mesh, \motion, \expos| V,B) = \frac{1}{N}\sum_{n=1}^{N} \| I_n  - \Ir_n(\mesh, \motion, \expos | B) \|_1 \enspace.
\end{equation}
This loss is low if the frames $\Ir_n$ rendered by our model via Eq.~\eqref{eq:msfb} closely look like the input frames $I_n$.

In order to make the optimization easier and well-behaved, we apply auxiliary loss terms and regularizers, similar to~\cite{sfb}. 
We briefly summarize them here and refer to~\cite{sfb} for details.
The silhouette consistency loss $\Loss_S$ helps localize the object in the image faster and serves as initialization for estimating the 3D mesh and its translation.
First, we run DeFMO~\cite{defmo} and use their estimated masks $\Mi$ for approximate object location.
To synchronize the motion direction (forward vs.~backward) for DeFMO masks across frames, we minimize the distance between consecutive masks in adjacent frames.
Then, $\Loss_S$ is defined as an intersection over union (IoU) between the DeFMO masks and 2D mesh silhouettes, rendered by our method: 
\begin{equation}
	\label{eq:sil}
	\begin{split}
	    \Loss_S = 
	     1 - \int_{0}^{1} \text{IoU} \Bigg(\Mi,  \Ren_S\bigg(\Mot\Big(\mesh, \TrajR(\Ti), \TrajT(\Ti)\Big)\bigg) \Bigg) \:\text{d}\Ti. \\
	\end{split}
\end{equation}

Furthermore, we add the commonly employed~\cite{pixel2mesh,dibr,sfb,tbd_ijcv,kotera2020} total variation and Laplacian regularizers.
Total variation $\Loss_T(\mesh)$ on texture maps encourages the model to produce smooth textures, and the Laplacian regularizer $\Loss_L(\mesh)$ promotes smooth meshes.
Finally, the joint loss is a weighted sum of all four loss terms:
\begin{equation}
	\label{eq:jloss}
    \begin{split}
	\Loss(\mesh,\motion,\expos | V, B) & = \lambda_V \cdot \Loss_V(\mesh,\motion,\expos  | V, B) +  \Loss_T(\mesh) + \\
	 &  + \Loss_S(\mesh,\motion,\expos | V, B) +  \lambda_L \cdot \Loss_L(\mesh) \enspace. \\
	\end{split}
\end{equation}

\newcommand{\AlgName}[2]{\multirow{1}{*}[#2]{\rotatebox{90}{\scriptsize #1 \phantom{Iy}}}}
\newcommand{\addimg}[1]{\includegraphics[angle=90,origin=c,width=0.077\linewidth]{#1}}
\newcommand{\makeRowSin}[1]{\addimg{#10} & \addimg{#11} & \addimg{#12} & \addimg{#13} &  \addimg{#14} & \addimg{#15} & \addimg{#16}  & \addimg{#17}}
\newcommand{\makeRow}[3]{%
\AlgName{Inputs}{35pt} & \addimg{#1_im} & & & \addimg{#2_im} & & \raisebox{0.2 em}{Bounce!} &  & & & & & \addimg{#3_im} & \\
\AlgName{DeFMO~\cite{defmo}}{47pt} & \addimg{#1_defmo3} & & \addimg{#1_defmo7}  & \makeRowSin{#2_defmo} & \addimg{#3_defmo0} & & \addimg{#3_defmo7}  \\
\AlgName{SfB~\cite{sfb}}{40pt} &  \addimg{#1_est3} &  $\dots$ &  \addimg{#1_est7}  & \makeRowSin{#2_est} & \addimg{#3_est0} & $\dots$ & \addimg{#3_est7}  \\
\AlgName{MfB (ours)}{45pt} & \addimg{#1_mfb3} & & \addimg{#1_mfb7}  & \makeRowSin{#2_mfb} & \addimg{#3_mfb0} & & \addimg{#3_mfb7}  \\
\AlgName{GT}{30pt} & \addimg{#1_hs3} & & \addimg{#1_hs7}  & \makeRowSin{#2_hs} & \addimg{#3_hs0} & & \addimg{#3_hs7} \\
 & \MakeArrowVar{3} & \MakeArrowVar{8} & \MakeArrowVar{3} \\\addlinespace[0.45em]
 & \multicolumn{3}{c}{$I_1$} & \multicolumn{8}{c}{$I_2$} & \multicolumn{3}{c}{$I_3$} \\
}

\begin{figure*}
\centering
\small
\setlength{\tabcolsep}{0.0em} 
\renewcommand{\arraystretch}{0} 
\begin{tabular}{@{}c@{\hskip 0.1em}ccc@{\hskip 0.1em}cccccccc@{\hskip 0.1em}ccc@{}}

\makeRow{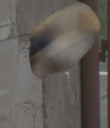}{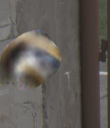}{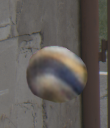}

\end{tabular}
\vspace{-7pt}
\caption{\textbf{Estimating 3D shape and motion of a motion-blurred volleyball, shown as temporal super-resolution.} 
The proposed Motion-from-Blur (MfB) method is the first to use multiple video frames during optimization and the first to model complex trajectories with bounces, accounting for the exposure gap. 
The previous methods for FMO deblurring (DeFMO) and single-frame 3D reconstruction (SfB) have difficulties reconstructing the bounce as they get confused by the ball's shadow due to the lack of multi-frame optimization.}
\label{fig:bounce}
\end{figure*}

\boldparagraph{Optimization.}
Fig.~\ref{fig:pipeline} shows an overview of the pipeline.
We backpropagate the joint loss up to the mesh $\mesh$, motion parameters $\motion$, and exposure gap~$\expos$. 
Optimization is done with ADAM\cite{adam} using a learning rate of $0.1$.
In the beginning, we run pre-optimization for at most 100 iterations with $\lambda_V = 0$, thus omitting the video reconstruction loss and texture map updates.
Pre-optimization stops when the silhouette loss $\Loss_S$ becomes $<0.3$, meaning that the mesh silhouettes have average IoU $>0.7$ with the DeFMO masks.
This pre-optimization phase is required since the 3D translation has to put the mesh at approximately the right location in the image to get a training signal for the video reconstruction loss $\Loss_V$ to estimate the texture map, 3D object rotation, and 3D shape.
The more video frames $N$ are used, the more important this step becomes because the object's 2D location varies more across the frames.
Experimentally, for $N > 2$ the optimization never converges without pre-optimization.
We optimize over the mesh prototypes by running the optimization for each prototype and choosing the best one based on the lowest value of the video reconstruction loss~\eqref{eq:loss}.
During optimization, the mesh is always kept in canonical space by normalizing the vertices to zero mean and unit variance. 
The main optimization is run for 1000 iterations using the full loss \eqref{eq:jloss} with $\lambda_V = 1$. 
The hyperparameter $\lambda_L$ of the Laplacian regularizer $\Loss_L$ is set to 1000 experimentally. 
Both the texture total variation $\Loss_T$ and silhouette consistency $\Loss_S$ losses have no weights since the default value of $1$ worked well in our experiments.
%

\boldparagraph{Initialization.}
The mesh parameters $\mesh$ are initialized to the prototype shape with zero vertex offsets and a white texture map.
The motion parameters $\motion$ are initialized such that the object is placed in the middle of the image with zero rotation.
Finally, the exposure gap $\expos$ is initialized to $0.1$.

\boldparagraph{Implementation.}
We use PyTorch~\cite{pytorch} with Kaolin~\cite{kaolin} for differentiable rendering.
All integrals in each frame are discretized by splitting time intervals into 8 evenly-spaced pieces.
All experiments are run on an Nvidia GTX 1080Ti GPU with $60$ seconds average runtime per frame.

\section{Experiments}  \label{sec:exper}
We evaluate our method's accuracy by measuring the deblurring quality on 3 real-world datasets from the fast moving object deblurring benchmark~\cite{defmo}.
Since there are no real image datasets of fast moving objects with associated ground-truth 3D shapes and motion, we follow the protocol of \cite{sfb} and evaluate the quality of reconstructed 3D meshes, 3D translations, and 3D rotations on a synthetic dataset.

\newcommand{\winner}[1]{\textbf{#1}}
\begin{table*}
\centering
\small
\setlength{\tabcolsep}{16.5pt} 
\newcommand{\MySkip}{\hskip 1.3em}  
\begin{tabular}{lc@{\MySkip}c@{\MySkip}cc@{\MySkip}c@{\MySkip}cc@{\MySkip}c@{\MySkip}c}
\toprule
\multirow{2}{*}{Method} & \multicolumn{3}{c}{Falling Objects~\cite{kotera2020}} & \multicolumn{3}{c}{TbD-3D Dataset~\cite{tbd3d}} & \multicolumn{3}{c}{TbD Dataset~\cite{tbd}}   \\
\cmidrule(lr){2-4} \cmidrule(lr){5-7} \cmidrule(lr){8-10} 
 & TIoU$\uparrow$ & PSNR$\uparrow$ & SSIM$\uparrow$ & TIoU$\uparrow$ & PSNR$\uparrow$ & SSIM$\uparrow$ & TIoU$\uparrow$ & PSNR$\uparrow$ & SSIM$\uparrow$ \\
\midrule
Jin~\etal~\cite{Jin_2018_CVPR} & N / A & 23.54 & 0.575 & N / A & 24.52 & 0.590 & N / A & 24.90 & 0.530 \\
DeblurGAN-v2~\cite{Kupyn_2019_ICCV} & N / A & 23.36 & 0.588 & N / A & 23.58 & 0.603 & N / A & 24.27 & 0.537 \\
TbD~\cite{tbd} & 0.539 & 20.53 & 0.591 & 0.598 & 18.84 & 0.504 & 0.542 & 23.22 & 0.605 \\
TbD-3D~\cite{tbd3d} & 0.539 & 23.42 & 0.671 & 0.598 & 23.13 & 0.651 & 0.542 & 25.21 & 0.674 \\
DeFMO~\cite{defmo} & 0.684 & 26.83 & 0.753 & 0.879 & 26.23 & 0.699 & 0.550 & 25.57 & 0.602 \\
SfB~\cite{sfb} & 0.701 & 27.18  & 0.760 & 0.921 & 26.54 & 0.722 & 0.610 & 25.66 &  0.659 \\ 
MfB (ours) & \winner{0.772} &  \winner{27.54} & \winner{0.765} & \winner{0.927} & \winner{26.57} & \winner{0.728} & \winner{0.614} & \winner{26.63} & \winner{0.678}  \\ 
\bottomrule
\end{tabular}
\vspace{-6pt}
\caption{\textbf{Fast moving object deblurring benchmark.} 
We compare the proposed MfB method to generic deblurring methods~\cite{Kupyn_2019_ICCV,Jin_2018_CVPR} (no trajectory output, thus TIoU is undefined) and to methods specifically designed for fast moving object deblurring~\cite{tbd,tbd3d,defmo,sfb}.}
\label{tab:datasets}
\vspace{-10pt}
\end{table*}
\newcommand{\BName}[2]{ \raisebox{0.4em}{\multirow{#1}{*}{\rotatebox[origin=l]{90}{\scriptsize #2 }}} }

\begin{table}
\centering
\small
\setlength{\tabcolsep}{16.5pt} 
\begin{tabular}{@{}l@{}lccc}
\toprule  
 & &  TIoU$\uparrow$ & PSNR$\uparrow$ & SSIM$\uparrow$ \\
\midrule
\BName{3}{full} 
 & SfB~\cite{sfb} & 0.921 & 26.54 & 0.722 \\ 
 & MfB (ours) &  \winner{0.927} & \winner{26.57} & \winner{0.728}  \\ 
\midrule
\BName{3}{bnc $\pm 1$} 
 & SfB~\cite{sfb} & 0.892 & 21.77 & 0.628 \\ 
 & MfB (ours) & \winner{0.902} & \winner{25.01} & \winner{0.643} \\ 
\midrule
\BName{3}{bnc $\pm 0$} 
 & SfB~\cite{sfb} & 0.863 & 20.77 & 0.595 \\ 
 & MfB (ours) &  \winner{0.889} & \winner{24.57} & \winner{0.620} \\ 
\bottomrule
\end{tabular}
\vspace{-6pt}
\caption{\textbf{Deblurring quality at bounces.} 
We compare scores on the full TbD-3D dataset~\cite{tbd3d}, on a subset of frames at bounces (bnc~$\pm 0$), and additionally on frames that are immediately before and after the bounce (bnc $\pm 1$).
The proposed multi-frame MfB is significantly more accurate at bounces (Fig.~\ref{fig:bounce}) than the single-frame SfB, especially on the deblurring metric PSNR.}
\label{tab:bounces}
\end{table}

\boldparagraph{Fast moving object deblurring benchmark.}
It consists of 3 datasets of varying difficulty.
The easiest one is TbD~\cite{tbd} that contains mostly spherical objects with uniform color (12 sequences, total 471 frames).
A more difficult dataset is TbD-3D~\cite{tbd3d} that contains mostly spherical objects with complex textures that move with significant 3D rotation (10 sequences, total 516 frames).
The most difficult dataset is Falling Objects~\cite{kotera2020} with objects of various shapes and complex textures (6 sequences, total 94 frames).
The ground truth for these datasets was recorded by a high-speed camera capturing the moving object without motion blur.
Therefore, we have 8 high-speed frames for each frame input to our method.
We measure the deblurring quality by reconstructing the high-speed camera footage as temporal super-resolution. 
For that, we apply the video formation model~\eqref{eq:msfb} at a 8 times finer temporal resolution by using the estimated object parameters after optimization on the input slow-speed frames.
Then, the reconstructed high-speed camera frames and the ground-truth ones are compared by the Peak Signal to Noise Ratio (PSNR) and Structural Similarity (SSIM) metrics.
Additionally, these datasets contain ground-truth 2D object trajectories and 2D object masks.
Therefore, we also measure the trajectory intersection over union (TIoU), defined as the IoU between the ground-truth mask placed at the ground-truth 2D location and the reconstructed 2D location (averaged over time).
We reconstruct the 2D object location for our method as the center of mass of the projected mesh silhouette at each high-speed frame.

We compare to various state-of-the-art methods:
a generic deblurring method DeblurGAN-v2~\cite{Kupyn_2019_ICCV},
a generic method for temporal super-resolution~\cite{Jin_2018_CVPR},
and methods designed for fast moving object deblurring~\cite{tbd,tbd3d,defmo,sfb}. 
All compared methods use each video frame independently, whereas MfB is the first method to exploit multiple frames simultaneously.
We run MfB in a temporal sliding window approach with $N=3$ if not mentioned otherwise.
For each frame, we always choose the window for which the video reconstruction loss~\eqref{eq:loss} is the lowest, measured only on this frame (similar to the best prototype selection).
This temporal sliding window approach reduces memory requirements and increases robustness to slightly moving camera and non-static background.


Table~\ref{tab:datasets} presents the results. 
MfB outperforms all other methods on all three datasets and for all three metrics.
Qualitatively, the estimated temporal super-resolution is more consistent compared to single-frame approaches since MfB explains all frames by a single 3D object mesh and texture (Fig.~\ref{fig:box}, $N=7$).
Novel view synthesis is also considerably better as the object outline is accurate from all viewpoints, and even sharp $90^{\circ}$ angles of the box (Fig.~\ref{fig:box}, novel views) are clear.
Interestingly, the previous state-of-the-art single-frame 3D reconstruction approach~\cite{sfb} produces several artifacts, inconsistencies, and produces an entirely incorrect 3D shape for object parts that are not visible in a single input frame.
Moreover, DeFMO~\cite{defmo} and SfB~\cite{sfb} fail in the presence of shadows and specularities, whereas MfB better reconstructs the object due to additional constraints from neighboring frames (Fig.~\ref{fig:box}, $I_1$ and $I_2$).

\boldparagraph{Exposure gap consistency.}
We evaluate the variance of the exposure gap estimation over the sequence duration, averaged over all sequences. The value is very low ($\sigma^2 = 0.002$), indicating good consistency.
Besides, the estimated exposure gap varies widely depending on the camera settings: $\tau = 0.05$ on the Falling Objects dataset, and $\tau = 0.7$ on the YouTube sequence in Fig.~\ref{fig:traj} (bottom).
This highlights the need for modeling the exposure gap.

\newcommand{\BSName}[2]{ \raisebox{0.1em}{\multirow{#1}{*}{\rotatebox[origin=l]{90}{\scriptsize #2 }}} }

\begin{table}
\centering
\small
\setlength{\tabcolsep}{16.5pt} 
\begin{tabular}{@{}l@{}lccc}
\toprule
 & & $\error_{\To}\!\downarrow$ & $\error_{\Ro}\!\downarrow$ & $\error_{\mesh}\!\downarrow$ \\
\midrule
\BSName{2}{$< 90 ^{\circ}$}
 & SfB~\cite{sfb} & 37.8 \% & 10.9$^{\circ}$ &  3.0 \% \\ 
 & MfB (ours) & \winner{20.0 \%} &  \winner{6.4$^{\circ}$} &   \winner{2.7 \%}  \\ 
\midrule
\BSName{2}{$< 30 ^{\circ}$}
 & SfB~\cite{sfb} & 12.8 \% & 4.8$^{\circ}$ & 2.3 \% \\ 
 & MfB (ours) & \winner{8.8 \%} & \winner{3.7$^{\circ}$} & \winner{2.2 \%}  \\ 
\bottomrule
\end{tabular}
\vspace{-6pt}
\caption{\textbf{Evaluating 3D translation, 3D rotation, and 3D shape on a synthetic dataset.} 
\textit{First block:} dataset with at most 90$^{\circ}$ rotation over 3 frames, \textit{second block:} at most 30$^{\circ}$ rotation.
The error rate of MfB is half of the single-frame SfB on the large rotation dataset when measuring 3D translation $\error_{\To}$ and 3D rotation $\error_{\Ro}$ errors. MfB is still significantly better on the small rotation dataset.}
\label{tab:synth}
\end{table}



\boldparagraph{Evaluating at bounces.}
A unique new feature of our approach is its ability to model bounces, which results in better deblurring in those cases.
Here, we evaluate this effect explicitly.
To this end, we manually annotate the frames in which a bounce happens in the TbD-3D dataset~\cite{tbd3d} (the only dataset with relatively frequent bounces).
Overall, we found 38 bounces from 516 frames in total from 10 sequences, which amounts to $7.4 \%$ chance of a bounce.
Since the frames immediately before and after a bounce are usually affected too (\eg due to a shadow as in Fig.~\ref{fig:bounce}), we also evaluate them, yielding a total of 114 frames ($22 \%$).
As shown in Table~\ref{tab:bounces}, MfB significantly outperforms SfB at bounces, especially in terms of deblurring quality metric PSNR.
The performance gap is still significant when evaluating on frames that are adjacent to the bounce but is relatively small when averaged over the whole dataset.
This indicates that bounces are significantly more difficult than other parts of the dataset, as shown qualitatively in Fig.~\ref{fig:traj} and Fig.~\ref{fig:bounce}, and our method successfully reconstructs such frames as well.
For single-frame approaches, the difficulty comes mainly from the trajectory non-linearity, slight object deformation, and shadows near the bounce point.
Motion-from-Blur is robust to these difficulties since the optimization is more constrained from easier frames before and after the bounce, and the trajectory is explicitly modeled with a bounce.
On frames that are far from the bounce, the difference in deblurring quality between the single-frame and multi-frame approaches is marginal on the TbD-3D dataset.
Note that our model is generic and estimates continuously connected trajectories also if there is no bounce.

\boldparagraph{Synthetic 3D dataset.}
We construct a synthetic dataset of fast moving objects with ground-truth 3D models and 3D motions for evaluation.
We sample random 3D models from the ShapeNet dataset~\cite{shapenet2015}, random linear 3D translations and 3D rotations (for a fair comparison with SfB~\cite{sfb} that reconstructs only linear motions), and random consecutive frames from the VOT~\cite{VOT_TPAMI} tracking dataset as backgrounds.
3D translation is randomly chosen in the interval between 1 to 5 object sizes, and 3D rotation is randomly chosen up to $30^{\circ}$ (first subset) or $90^{\circ}$ (second subset) during the video duration. 
Then, we apply the video formation model~\eqref{eq:msfb} with $N=3$ to create two subsets, each consisting of 30 short videos.
We report the mesh error $\error_{\mesh}$ as the average bidirectional distance between the closest vertices of the ground-truth and the estimated mesh, both placed at the ground-truth and predicted initial 6D pose, and divided by the object size. 
For evaluating the translation error $\error_{\To}$, we compute the norm of the difference vector between the predicted and ground-truth translation offset $\TrajT(1) - \TrajT(0)$, divided by the object size.
Thus, these two scores ($\error_{\mesh}$ and $\error_{\To}$) are reported as a fraction of the object size. 
For evaluating the rotation error $\error_{\Ro}$, we compute the average angle between the estimated rotation change (rotation between $\TrajR(1)$ and $\TrajR(0)$) and the ground-truth one. 

We compare to the only other method that can reconstruct a 3D object and its motion from the motion-blurred input (SfB~\cite{sfb}).
Our method is applied to all three video frames in each video, whereas SfB is applied to them individually, and the scores are averaged (w.r.t.~one video frame).
As shown in Table~\ref{tab:synth}, on the synthetic dataset~\footnote{All figures show only real data.} with up to $90^{\circ}$ rotation, our method is almost twice as accurate as SfB in terms of 3D translation and 3D rotation estimation.
For smaller rotations up to $30^{\circ}$, the difference is smaller but is still significant.
This highlights that multi-frame optimization is especially beneficial for complex objects (as from ShapeNet) with non-negligible rotations.

\boldparagraph{Applications.}
MfB can be used for imitating high-speed cameras or multiplying their capabilities by creating temporal super-resolution from motion-blurred videos.
MfB can perform 3D reconstruction of blurred objects that are almost unidentifiable by humans,~\eg image forensics of surveillance cameras.
Applications also include 6D object tracking and reconstruction in sports,~\eg football, tennis, basketball.

\newcommand{\addimgTraj}[1]{\includegraphics[width=0.16\linewidth]{imgs/traj/#1}}

\begin{figure}
\centering
\small
\setlength{\tabcolsep}{0.05em} 
\renewcommand{\arraystretch}{0.4} 
\begin{tabular}{@{}ccccccc@{}}
\addimgTraj{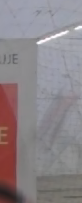} &
\addimgTraj{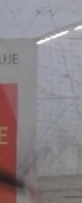} &
\addimgTraj{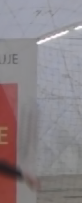} &
\addimgTraj{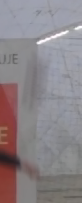} &
\addimgTraj{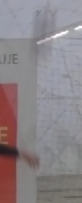} & 
\addimgTraj{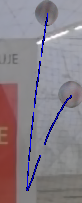} \\

\addimgTraj{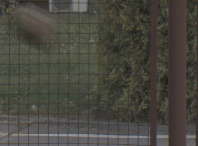} &
\addimgTraj{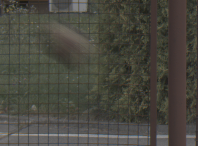} &
\addimgTraj{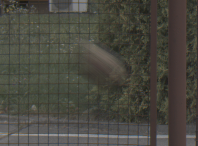} &
\addimgTraj{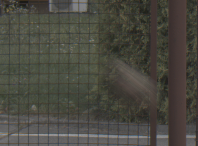} &
\addimgTraj{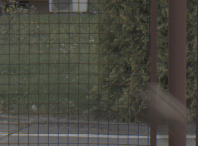} & 
\addimgTraj{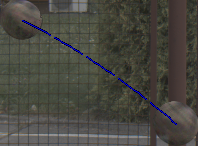} \\

\addimgTraj{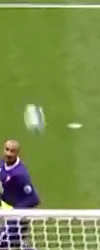} &
\addimgTraj{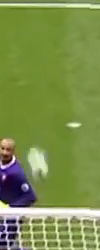} &
\addimgTraj{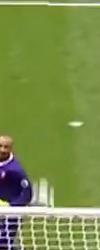} &
\addimgTraj{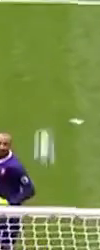} &
 \raisebox{5em}{$\dots$} &
\addimgTraj{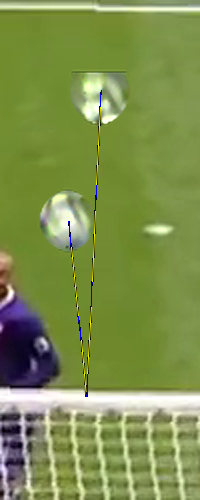} \\

$I_1$ & $I_2$ & $I_3$ & $I_4$ & $I_5$ & Traj. \\

\end{tabular}
\vspace{-8pt}
\caption{\textbf{Reconstructing 2D object trajectories with bounces.} For each video, we reconstruct 3D object and its motion (\textit{blue}: observed trajectory, \textit{yellow}: the exposure gap).
We visualize the trajectory for the center of mass of the mesh silhouettes and further render the first and last pose of the object (right-most image).
\textbf{Top row:} Scene from TbD~\cite{tbd} dataset; \textbf{Center row:} TbD-3D~\cite{tbd3d} scene; \textbf{Bottom row:} \href{https://youtu.be/wa1ORli2Nyo?t=78}{YouTube} scene from Fig.~\ref{fig:pipeline}.}
\label{fig:traj}
\end{figure}

\newcommand{\addimgtext}[2]{\addimgB{#1}%
\raisebox{3pt}{\makebox[0pt][r]{#2 \phantom{aa} }}}

\newcommand{\addimgB}[1]{\includegraphics[width=0.05\linewidth]{#1}}
\newcommand{\makeOneBlock}[1]{\addimgB{#10}&\addimgB{#17}}

\newcommand{\makeRowBoxSeven}[7]{%
\AlgName{Inputs}{35pt} & 
\addimgB{#1_im} & & 
\addimgB{#2_im} & & 
\addimgB{#3_im} & &   
\addimgB{#4_im} & & 
\addimgB{#5_im} & & \addimgB{#6_im} & &\addimgB{#7_im} & & \\
\AlgName{DeFMO~\cite{defmo}}{55pt} & 
\makeOneBlock{#1_defmo} & \makeOneBlock{#2_defmo} & \makeOneBlock{#3_defmo} & \makeOneBlock{#4_defmo}  & \makeOneBlock{#5_defmo} & \makeOneBlock{#6_defmo} & \makeOneBlock{#7_defmo} & \multicolumn{3}{c}{\smash{\raisebox{4em}{No, cannot do.}}}\\
\AlgName{SfB~\cite{sfb}}{55pt} &  
\makeOneBlock{#1_sfb} & \makeOneBlock{#2_sfb} & \makeOneBlock{#3_sfb} & \makeOneBlock{#4_sfb}  & \makeOneBlock{#5_sfb} & \makeOneBlock{#6_sfb} & \makeOneBlock{#7_sfb} & \addimgtext{#1_novel}{$I_1$} & \addimgtext{#3_novel}{$I_3$} & \addimgtext{#5_novel}{$I_5$}  \\
\AlgName{MfB (ours)}{55pt} & \makeOneBlock{#1_mfb} & \makeOneBlock{#2_mfb} & \makeOneBlock{#3_mfb} & \makeOneBlock{#4_mfb}  & \makeOneBlock{#5_mfb} &\makeOneBlock{#6_mfb} &\makeOneBlock{#7_mfb} & \addimgB{#1_novel_mfb1} & \addimgB{#1_novel_mfb2} & \addimgB{#1_novel_mfb3} \\
\AlgName{GT}{40pt} & 
\makeOneBlock{#1_hs} & \makeOneBlock{#2_hs} & \makeOneBlock{#3_hs} & \makeOneBlock{#4_hs}  & \makeOneBlock{#5_hs}& \makeOneBlock{#6_hs}& \makeOneBlock{#7_hs} &  \multicolumn{3}{c}{\smash{\raisebox{4em}{Not available.}}} \\
}
\newcommand{\MySkip}{\hskip 0.4em}
\begin{figure*}
\centering
\small
\setlength{\tabcolsep}{0.05em} 
\renewcommand{\arraystretch}{0.3} 
\begin{tabular}{@{}c@{\hskip 0.1em}cc@{\MySkip}cc@{\MySkip}cc@{\MySkip}cc@{\MySkip}cc@{\MySkip}cc@{\MySkip}cc@{\MySkip}ccc@{}}

\makeRowBoxSeven{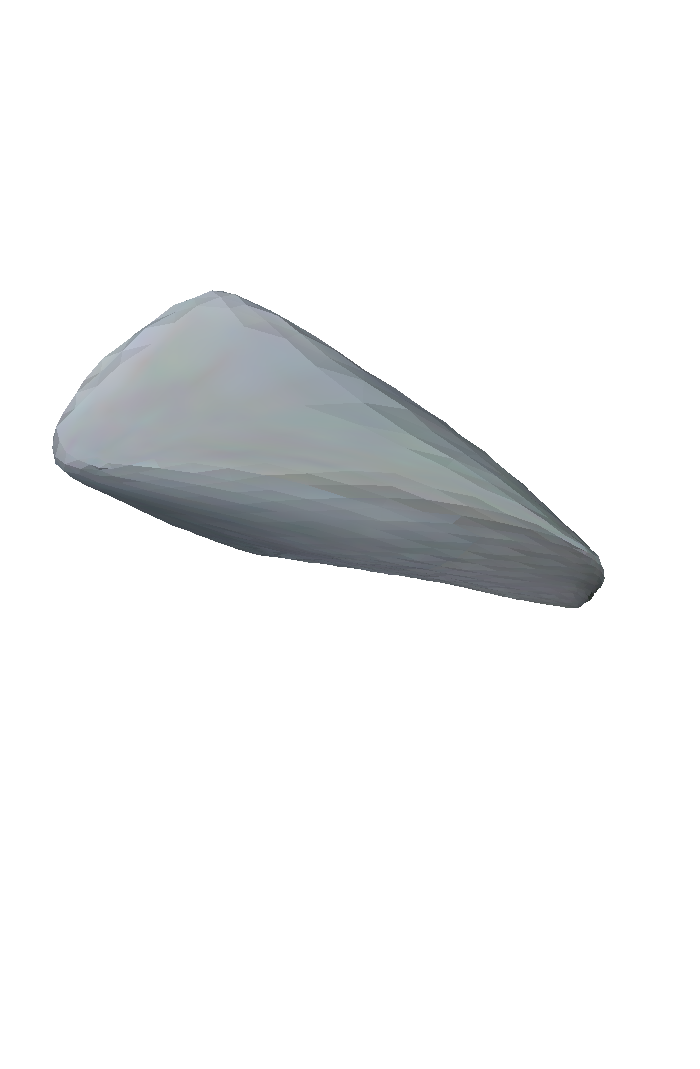}{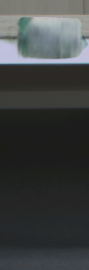}{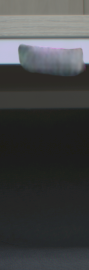}{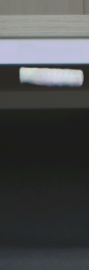}{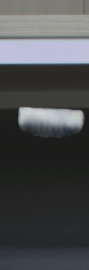}{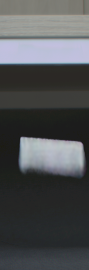}{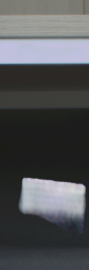}
 & \MakeArrowTwo & \MakeArrowTwo & \MakeArrowTwo & \MakeArrowTwo & \MakeArrowTwo & \MakeArrowTwo & \MakeArrowTwo \\
 \addlinespace[0.4em]
 & \multicolumn{2}{c}{$I_1$} & \multicolumn{2}{c}{$I_2$} & \multicolumn{2}{c}{$I_3$} & \multicolumn{2}{c}{$I_4$}  & \multicolumn{2}{c}{$I_5$}  & \multicolumn{2}{c}{$I_6$} & \multicolumn{2}{c}{$I_7$}   & \multicolumn{3}{c}{Novel views} \\
\end{tabular}
\vspace{-8pt}
\caption{\textbf{3D reconstruction and temporal super-resolution of a falling box from Falling Objects dataset~\cite{kotera2020}.} Our method produces more consistent results over $N = 7$ input frames than previous methods and does not suffer from artifacts on frames with shadows ($I_1$ and $I_2$). 
The final 3D reconstruction is also more complete and accurate than the single-frame approach SfB~\cite{sfb} as shown on novel views.}
\label{fig:box}
\end{figure*}

\section{Limitations}  \label{sec:limit}
%
\boldparagraph{Static camera.}
MfB assumes that the video is captured by a nearly static camera.
A moving camera adds even more ambiguity to the observed blur that could stem from both camera and object motion blur.
Moreover, motion blur also has to be compensated by the camera motion, and the whole problem would become much more difficult.
Since all previous methods for fast moving object deblurring and 3D reconstruction~\cite{tbd,tbd3d,tbd_ijcv,defmo,sfb} also assume a static camera, tackling this problem remains challenging future work.

\boldparagraph{Shutter.}
We assume that the shutter speed is constant.
However, some cameras have an adjustable shutter that changes the exposure gap based on lighting conditions,~\eg less exposure for bright scenes and more exposure for dark scenes.
In most cases, this transition is smooth, and our method is robust thanks to the sliding window approach.
%
Modeling a rolling shutter is beyond the scope of this paper.

\boldparagraph{Texture-less objects.}
Reconstructing 3D objects that lack noticeable texture is a challenge even for generic 3D reconstruction since no distinctive geometry features are observable, and the correspondences are ambiguous.
In this case, detecting any 3D rotation is almost infeasible.

\boldparagraph{Non-rigid objects.}
We assume that the object is rigid,~\ie its 3D model is constant for the video duration. 
Such assumption is invalid for deforming objects, which often happens during the bounce.
However, since these deformations are often insignificant and only for a very short duration of time, our modeling still handles such cases well.

\blfootnote{\textbf{Acknowledgements.} This research was supported by a Google Focused Research Award, Innosuisse grant No.~34475.1 IP-ICT, Research Center for Informatics (project CZ.02.1.01/0.0/0.0/16$\_$019/0000765 funded by OP VVV), and a research grant by FIFA.}

\vspace{-27pt}
\section{Conclusion}
We presented the first method for estimating textured 3D shapes and complex motions of motion-blurred objects in videos.
By optimizing over multiple input frames, we correctly recover 3D object shape and motion, its motion direction, and the camera exposure gap.
Various experiments have shown that our method produces sharper and more consistent results compared to other methods for fast moving object deblurring.
Compared to single-image 3D shape and motion estimation~\cite{sfb}, which is a special instance of our approach, we recover more complete shapes and significantly more precise motion estimation.

\section*{Appendix}
\appendix
\section{Evaluation on foreground regions}
In the main paper, PSNR and SSIM metrics are calculated over the tightly cropped input frame.
However, we can also evaluate on the foreground regions defined by the object mask in the ground-truth, which occupies even less area.
In Tab.~\ref{tab:mask} we now evaluate SfB and our MfB only on the ground-truth foreground region.
The improvement made by MfB over SfB is larger but the overall scores are lower than when evaluating on the cropped frame.
This is expected since background regions are easier as they are static.

\section{Synthetic noise}
We run an experiment with additive Gaussian noise, ranging from small to extremely high standard deviation (std).
To quantify degradation, in Fig.~\ref{fig:noise} we measure the same performance metrics and divide them by the original ones without noise (on the Falling Objects and TbD-3D datasets).
The method appears robust, \eg when adding noise at 5\% of the full RGB range, performance degrades by only 1\%-4\% depending on the metric on the TbD-3D dataset.

\section{Generalization ability}
Based on the experiments, we observed that our method can handle various shapes such as a pen, a key, and an Aerobie ring. 
Since we deform a set of mesh prototypes, our method is not able to generalize to other topologies.
Therefore, objects can be quite general once they have the same topology as in the prototype set. Shapes that are too difficult to handle are very spiky objects,~\eg a chair with four legs. 
However, even in such cases, we can add more prototypes, which will cover more object classes and make the method more general. 
We leave this for future work.

\section{Video processing}
The YouTube video, as all other videos, was processed fully automatically.
In each frame, we run a state-of-the-art fast moving object detection method, which outputs a bounding box.
Outside the bounding box, the input video is not used and is kept unchanged. 
In the cropped region, we assume that the camera and the background are stationary. 

\begin{table}
\centering
\small
\setlength{\tabcolsep}{8.5pt} 
\begin{tabular}{@{}lccccc}
\toprule  
 & \multicolumn{2}{c}{Falling Objects} & \multicolumn{2}{c}{TbD-3D Dataset} \\
 \cmidrule(lr){2-3}   \cmidrule(lr){4-5}
 & PSNR$\uparrow$ & SSIM$\uparrow$ & PSNR$\uparrow$ & SSIM$\uparrow$  \\
\midrule
SfB &  21.77 & 0.597  & 21.16 & 0.528 \\ 
MfB (ours) &   \winner{23.43} & \winner{0.649} &  \winner{22.91} & \winner{0.612}   \\ 
\bottomrule
\end{tabular}
\vspace{-6pt}
\caption{\textbf{Evaluation on foreground regions only.}}
\vspace{-6pt}
\label{tab:mask}
\end{table}




\newcommand{\myPlot}[1]{\resizebox{!}{0.33\linewidth}{\begin{tikzpicture}
\begin{axis}[xlabel={Standard deviation of Gaussian noise [$\%$]},ylabel={Score degradation},yticklabel style={/pgf/number format/fixed, /pgf/number format/precision=2},scaled y ticks=false, xticklabel style={/pgf/number format/fixed, /pgf/number format/precision=1},width=0.45\textwidth,height=0.32\textwidth,legend style={at={(0,0)},anchor=south west}]
\addplot table[x=noise, y=TIoU#1, col sep=comma] {noise.dat};
\addlegendentry{TIoU}
\addplot table[x=noise, y=PSNR#1, col sep=comma] {noise.dat};
\addlegendentry{PSNR}
\addplot table[x=noise, y=SSIM#1, col sep=comma] {noise.dat};
\addlegendentry{SSIM}
\end{axis}
\end{tikzpicture}}}

\begin{figure*}
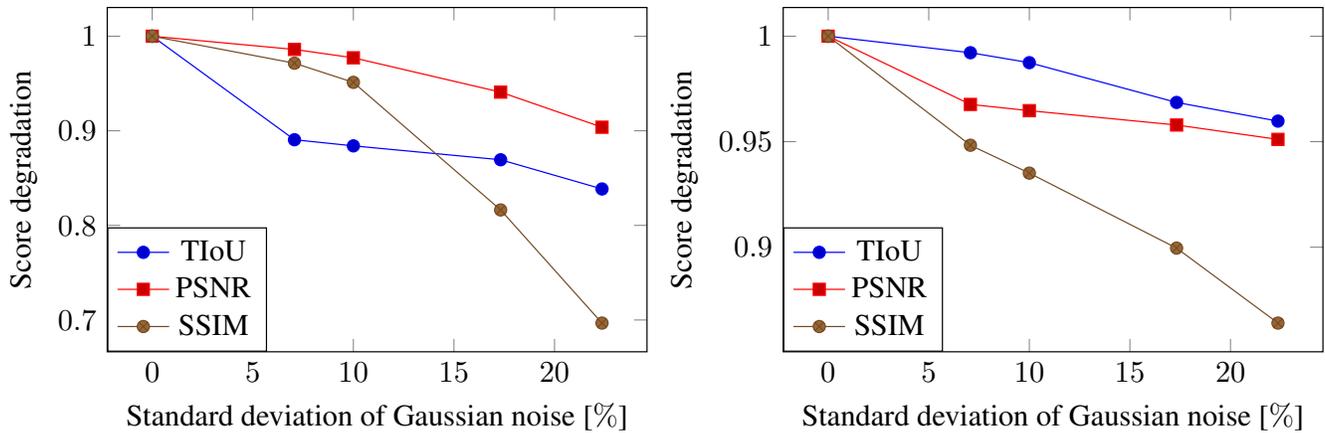

\centering
\begin{tabular}{@{}c@{\hskip 0.5em}c@{}}
\myPlot{} & \myPlot{1} \\ 
\end{tabular}
\vspace{-12pt}
\caption{\textbf{Score degradation with increasing Gaussian noise.} Left: Falling Objects dataset. Right: TbD-3D dataset.}
\vspace{-15pt}
\label{fig:noise}
\end{figure*}

{\small
\bibliographystyle{ieee_fullname}
\bibliography{egbib}

\begin{thebibliography}{10}\itemsep=-1pt

\bibitem{shapenet2015}
Angel~X. Chang, Thomas Funkhouser, Leonidas Guibas, Pat Hanrahan, Qixing Huang,
  Zimo Li, Silvio Savarese, Manolis Savva, Shuran Song, Hao Su, Jianxiong Xiao,
  Li Yi, and Fisher Yu.
\newblock {ShapeNet: An Information-Rich 3D Model Repository}.
\newblock Technical Report arXiv:1512.03012 [cs.GR], Stanford University ---
  Princeton University --- Toyota Technological Institute at Chicago, 2015.

\bibitem{dibr}
Wenzheng Chen, Jun Gao, Huan Ling, Edward Smith, Jaakko Lehtinen, Alec
  Jacobson, and Sanja Fidler.
\newblock Learning to predict 3d objects with an interpolation-based
  differentiable renderer.
\newblock In {\em NeurIPS}, 2019.

\bibitem{Chi_2021_CVPR}
Zhixiang Chi, Yang Wang, Yuanhao Yu, and Jin Tang.
\newblock Test-time fast adaptation for dynamic scene deblurring via
  meta-auxiliary learning.
\newblock In {\em Proceedings of the IEEE/CVF Conference on Computer Vision and
  Pattern Recognition (CVPR)}, pages 9137--9146, June 2021.

\bibitem{Ding_2021_CVPR}
Tianyu Ding, Luming Liang, Zhihui Zhu, and Ilya Zharkov.
\newblock Cdfi: Compression-driven network design for frame interpolation.
\newblock In {\em Proceedings of the IEEE/CVF Conference on Computer Vision and
  Pattern Recognition (CVPR)}, pages 8001--8011, June 2021.

\bibitem{Fan_2017_CVPR}
Haoqiang Fan, Hao Su, and Leonidas~J. Guibas.
\newblock A point set generation network for 3d object reconstruction from a
  single image.
\newblock In {\em CVPR}, July 2017.

\bibitem{Gui_2020_CVPR}
Shurui Gui, Chaoyue Wang, Qihua Chen, and Dacheng Tao.
\newblock Featureflow: Robust video interpolation via structure-to-texture
  generation.
\newblock In {\em IEEE/CVF Conference on Computer Vision and Pattern
  Recognition (CVPR)}, June 2020.

\bibitem{han2021reconstructing}
Muzhi Han, Zeyu Zhang, Ziyuan Jiao, Xu Xie, Yixin Zhu, Song-Chun Zhu, and
  Hangxin Liu.
\newblock Reconstructing interactive 3d scenes by panoptic mapping and cad
  model alignments.
\newblock In {\em ICRA}, 2021.

\bibitem{kaolin}
Krishna~Murthy Jatavallabhula, Edward Smith, Jean-Francois Lafleche,
  Clement~Fuji Tsang, Artem Rozantsev, Wenzheng Chen, Tommy Xiang, Rev
  Lebaredian, and Sanja Fidler.
\newblock Kaolin: A pytorch library for accelerating 3d deep learning research.
\newblock {\em arXiv:1911.05063}, 2019.

\bibitem{Jin_2019_CVPR}
Meiguang Jin, Zhe Hu, and Paolo Favaro.
\newblock Learning to extract flawless slow motion from blurry videos.
\newblock In {\em CVPR}, June 2019.

\bibitem{Jin_2018_CVPR}
Meiguang Jin, Givi Meishvili, and Paolo Favaro.
\newblock Learning to extract a video sequence from a single motion-blurred
  image.
\newblock In {\em CVPR}, June 2018.

\bibitem{Kaufman_2020_CVPR}
Adam Kaufman and Raanan Fattal.
\newblock Deblurring using analysis-synthesis networks pair.
\newblock In {\em IEEE/CVF Conference on Computer Vision and Pattern
  Recognition (CVPR)}, June 2020.

\bibitem{adam}
Diederik~P. Kingma and Jimmy Ba.
\newblock Adam: {A} method for stochastic optimization.
\newblock In Yoshua Bengio and Yann LeCun, editors, {\em ICLR}, 2015.

\bibitem{kotera2020}
J. {Kotera}, J. {Matas}, and F. {{\v{S}}roubek}.
\newblock Restoration of fast moving objects.
\newblock {\em IEEE TIP}, 29:8577--8589, 2020.

\bibitem{tbd}
J. Kotera, D. Rozumnyi, F. {\v{S}}roubek, and J. Matas.
\newblock Intra-frame object tracking by deblatting.
\newblock In {\em ICCVW}, Oct 2019.

\bibitem{kotera2018}
J. {Kotera} and F. {\v{S}roubek}.
\newblock Motion estimation and deblurring of fast moving objects.
\newblock In {\em ICIP}, pages 2860--2864, Oct 2018.

\bibitem{VOT_TPAMI}
Matej Kristan, Jiri Matas, Ale\v{s} Leonardis, Tomas Vojir, Roman Pflugfelder,
  Gustavo Fernandez, Georg Nebehay, Fatih Porikli, and Luka \v{C}ehovin.
\newblock A novel performance evaluation methodology for single-target
  trackers.
\newblock {\em IEEE TPAMI}, 38(11):2137--2155, Nov 2016.

\bibitem{Kupyn_2018_CVPR}
Orest Kupyn, Volodymyr Budzan, Mykola Mykhailych, Dmytro Mishkin, and Jiří
  Matas.
\newblock Deblurgan: Blind motion deblurring using conditional adversarial
  networks.
\newblock In {\em Proceedings of the IEEE Conference on Computer Vision and
  Pattern Recognition (CVPR)}, June 2018.

\bibitem{Kupyn_2019_ICCV}
Orest Kupyn, Tetiana Martyniuk, Junru Wu, and Zhangyang Wang.
\newblock Deblurgan-v2: Deblurring (orders-of-magnitude) faster and better.
\newblock In {\em ICCV}, Oct 2019.

\bibitem{Li_2021_CVPR}
Dongxu Li, Chenchen Xu, Kaihao Zhang, Xin Yu, Yiran Zhong, Wenqi Ren, Hanna
  Suominen, and Hongdong Li.
\newblock Arvo: Learning all-range volumetric correspondence for video
  deblurring.
\newblock In {\em Proceedings of the IEEE/CVF Conference on Computer Vision and
  Pattern Recognition (CVPR)}, pages 7721--7731, June 2021.

\bibitem{nerf}
Ben Mildenhall, Pratul~P. Srinivasan, Matthew Tancik, Jonathan~T. Barron, Ravi
  Ramamoorthi, and Ren Ng.
\newblock Nerf: Representing scenes as neural radiance fields for view
  synthesis.
\newblock In {\em ECCV}, 2020.

\bibitem{Niklaus_CVPR_2020}
Simon Niklaus and Feng Liu.
\newblock Softmax splatting for video frame interpolation.
\newblock In {\em CVPR}, 2020.

\bibitem{Pan_2020_CVPR}
Jinshan Pan, Haoran Bai, and Jinhui Tang.
\newblock Cascaded deep video deblurring using temporal sharpness prior.
\newblock In {\em CVPR}, June 2020.

\bibitem{pytorch}
Adam Paszke, Sam Gross, Francisco Massa, Adam Lerer, James Bradbury, Gregory
  Chanan, Trevor Killeen, Zeming Lin, Natalia Gimelshein, Luca Antiga, Alban
  Desmaison, Andreas Kopf, Edward Yang, Zachary DeVito, Martin Raison, Alykhan
  Tejani, Sasank Chilamkurthy, Benoit Steiner, Lu Fang, Junjie Bai, and Soumith
  Chintala.
\newblock Pytorch: An imperative style, high-performance deep learning library.
\newblock In H. Wallach, H. Larochelle, A. Beygelzimer, F. d Alch\'{e}-Buc, E.
  Fox, and R. Garnett, editors, {\em Advances in Neural Information Processing
  Systems 32}, pages 8024--8035. Curran Associates, Inc., 2019.

\bibitem{Richter_2018_CVPR}
Stephan~R. Richter and Stefan Roth.
\newblock Matryoshka networks: Predicting 3d geometry via nested shape layers.
\newblock In {\em CVPR}, June 2018.

\bibitem{tbdnc}
D. Rozumnyi, J. Kotera, F. {\v{S}}roubek, and J. Matas.
\newblock Non-causal tracking by deblatting.
\newblock In Gernot~A. Fink, Simone Frintrop, and Xiaoyi Jiang, editors, {\em
  GCPR}, pages 122--135, Cham, 2019. Springer International Publishing.

\bibitem{tbd_ijcv}
D. Rozumnyi, J. Kotera, F. {\v{S}}roubek, and J. Matas.
\newblock Tracking by deblatting.
\newblock {\em IJCV}, 129(9):2583–2604, 2021.

\bibitem{fmo}
D. {Rozumnyi}, J. {Kotera}, F. {\v{S}roubek}, L. {Novotn\'{y}}, and J. {Matas}.
\newblock The world of fast moving objects.
\newblock In {\em CVPR}, pages 4838--4846, July 2017.

\bibitem{tbd3d}
D. {Rozumnyi}, J. {Kotera}, F. {{\v{S}}roubek}, and J. {Matas}.
\newblock Sub-frame appearance and 6d pose estimation of fast moving objects.
\newblock In {\em CVPR}, pages 6777--6785, 2020.

\bibitem{fmodetect}
Denys Rozumnyi, Ji\v{r}{\'\i} Matas, Filip \v{S}roubek, Marc Pollefeys, and
  Martin~R. Oswald.
\newblock Fmodetect: Robust detection of fast moving objects.
\newblock In {\em ICCV}, pages 3541--3549, October 2021.

\bibitem{defmo}
Denys Rozumnyi, Martin~R. Oswald, Vittorio Ferrari, Jiri Matas, and Marc
  Pollefeys.
\newblock Defmo: Deblurring and shape recovery of fast moving objects.
\newblock In {\em CVPR}, Nashville, Tennessee, USA, Jun 2021.

\bibitem{sfb}
Denys Rozumnyi, Martin~R. Oswald, Vittorio Ferrari, and Marc Pollefeys.
\newblock Shape from blur: Recovering textured 3d shape and motion of fast
  moving objects.
\newblock In {\em NeurIPS}, 2021.

\bibitem{schoenberger2016sfm}
Johannes~Lutz Sch\"{o}nberger and Jan-Michael Frahm.
\newblock Structure-from-motion revisited.
\newblock In {\em Conference on Computer Vision and Pattern Recognition
  (CVPR)}, 2016.

\bibitem{schoenberger2016mvs}
Johannes~Lutz Sch\"{o}nberger, Enliang Zheng, Marc Pollefeys, and Jan-Michael
  Frahm.
\newblock Pixelwise view selection for unstructured multi-view stereo.
\newblock In {\em European Conference on Computer Vision (ECCV)}, 2016.

\bibitem{Shen_2020_CVPR}
Wang Shen, Wenbo Bao, Guangtao Zhai, Li Chen, Xiongkuo Min, and Zhiyong Gao.
\newblock Blurry video frame interpolation.
\newblock In {\em CVPR}, June 2020.

\bibitem{Siyao_2021_CVPR}
Li Siyao, Shiyu Zhao, Weijiang Yu, Wenxiu Sun, Dimitris Metaxas, Chen~Change
  Loy, and Ziwei Liu.
\newblock Deep animation video interpolation in the wild.
\newblock In {\em Proceedings of the IEEE/CVF Conference on Computer Vision and
  Pattern Recognition (CVPR)}, pages 6587--6595, June 2021.

\bibitem{sfm}
W{\l}adys{\l}aw Skarbek.
\newblock Shape from motion revisited.
\newblock In Dominik {\'{S}}lezak, Gerald Schaefer, Son~T. Vuong, and Yoo-Sung
  Kim, editors, {\em Active Media Technology}, pages 383--394, Cham, 2014.
  Springer International Publishing.

\bibitem{Stuhmer_2015_ICCV}
Jan Stuhmer, Sebastian Nowozin, Andrew Fitzgibbon, Richard Szeliski, Travis
  Perry, Sunil Acharya, Daniel Cremers, and Jamie Shotton.
\newblock Model-based tracking at 300hz using raw time-of-flight observations.
\newblock In {\em Proceedings of the IEEE International Conference on Computer
  Vision (ICCV)}, December 2015.

\bibitem{Suin_2020_CVPR}
Maitreya Suin, Kuldeep Purohit, and A.~N. Rajagopalan.
\newblock Spatially-attentive patch-hierarchical network for adaptive motion
  deblurring.
\newblock In {\em IEEE/CVF Conference on Computer Vision and Pattern
  Recognition (CVPR)}, June 2020.

\bibitem{what3d_cvpr19}
Maxim Tatarchenko*, Stephan~R. Richter*, René Ranftl, Zhuwen Li, Vladlen
  Koltun, and Thomas Brox.
\newblock What do single-view 3d reconstruction networks learn?
\newblock In {\em CVPR}, 2019.

\bibitem{sroubek2020}
F. {{\v{S}}roubek} and J. {Kotera}.
\newblock Motion blur prior.
\newblock In {\em ICIP}, pages 928--932, 2020.

\bibitem{pixel2mesh}
Nanyang Wang, Yinda Zhang, Zhuwen Li, Yanwei Fu, Wei Liu, and Yu-Gang Jiang.
\newblock Pixel2mesh: Generating 3d mesh models from single rgb images.
\newblock In {\em ECCV}, 2018.

\bibitem{Xu_2018_CVPR_Workshops}
Yi Xu, Yuzhang Wu, and Hui Zhou.
\newblock Multi-scale voxel hashing and efficient 3d representation for mobile
  augmented reality.
\newblock In {\em CVPRW}, June 2018.

\bibitem{pixelnerf}
Alex Yu, Vickie Ye, Matthew Tancik, and Angjoo Kanazawa.
\newblock {pixelNeRF}: Neural radiance fields from one or few images.
\newblock In {\em CVPR}, 2021.

\bibitem{Zhang_2020_CVPR}
Kaihao Zhang, Wenhan Luo, Yiran Zhong, Lin Ma, Bjorn Stenger, Wei Liu, and
  Hongdong Li.
\newblock Deblurring by realistic blurring.
\newblock In {\em CVPR}, June 2020.

\bibitem{fmo_segmentation}
Ale{\v{s}} Zita and Filip {\v{S}}roubek.
\newblock Tracking fast moving objects by segmentation network.
\newblock In {\em ICPR}, pages 10312--10319, 2021.

\end{thebibliography}
}

\end{document}